\documentclass[10pt,twocolumn,letterpaper]{article}

\usepackage{cvpr}
\usepackage{times}
\usepackage{epsfig}
\usepackage{graphicx}
\usepackage{amsmath}
\usepackage{amssymb}

\usepackage{booktabs}
\usepackage{multirow}
\usepackage{authblk}

\usepackage[breaklinks=true,bookmarks=false]{hyperref}

\cvprfinalcopy 

\def\cvprPaperID{****} 

\ifcvprfinal\fi
\begin{document}

\title{Parsing-based View-aware Embedding Network for Vehicle Re-Identification}

\author[1,2]{Dechao Meng}
\author[1]{Liang Li\thanks{Corresponding author.}}
\author[1,2]{Xuejing Liu}
\author[3]{Yadong Li}
\author[2]{Shijie Yang}
\author[4]{Zheng-Jun Zha}
\author[6]{Xingyu Gao}
\author[1]{Shuhui Wang}
\author[2,1,5]{Qingming Huang}
\affil[1]{Key Lab of Intell. Info. Process., Inst. of Comput. Tech., CAS, Beijing, China} 
\affil[2]{University of Chinese Academy of Sciences, China, $^3$Megvii Inc, Beijing, China}
\affil[4]{University of Science and Technology of China, China, $^5$Peng Cheng Laboratory, Shenzhen, China, }
\affil[6]{ Institute of Microelectronics, Chinese Academy of Sciences, Beijing, China}
\affil[ ]{\textit {{\{dechao.meng, xuejing.liu, shijie.yang\}@vipl.ict.ac.cn, \{liang.li, wangshuhui\}@ict.ac.cn, liyadong@megvii.inc, gaoxingyu@ime.ac.cn,  zhazj@ustc.edu.cn, qmhuang@ucas.ac.cn}}
}

\maketitle

    \begin{abstract}
Vehicle Re-Identification is to find images of the same vehicle from various views in the cross-camera scenario.
The main challenges of this task are the large intra-instance distance caused by different views and the subtle inter-instance discrepancy caused by similar vehicles.
In this paper, we propose a parsing-based view-aware embedding network (PVEN) to achieve the view-aware feature alignment and enhancement for vehicle ReID.
First, we introduce a parsing network to parse a vehicle into four different views, and then align the features by mask average pooling.
Such alignment provides a fine-grained representation of the vehicle.
Second, in order to enhance the view-aware features,
we design a common-visible attention to focus on the common visible views,
which not only shortens the distance among intra-instances, but also enlarges the discrepancy of inter-instances.
The PVEN helps capture the stable discriminative information of vehicle under different views.
The experiments conducted on three datasets show that our model outperforms state-of-the-art methods by a large margin.
\thispagestyle{empty}
    \end{abstract}

    \section{Introduction}\label{sec:introduction}
\begin{figure}[t]
    \centering
    \includegraphics[width=\columnwidth]{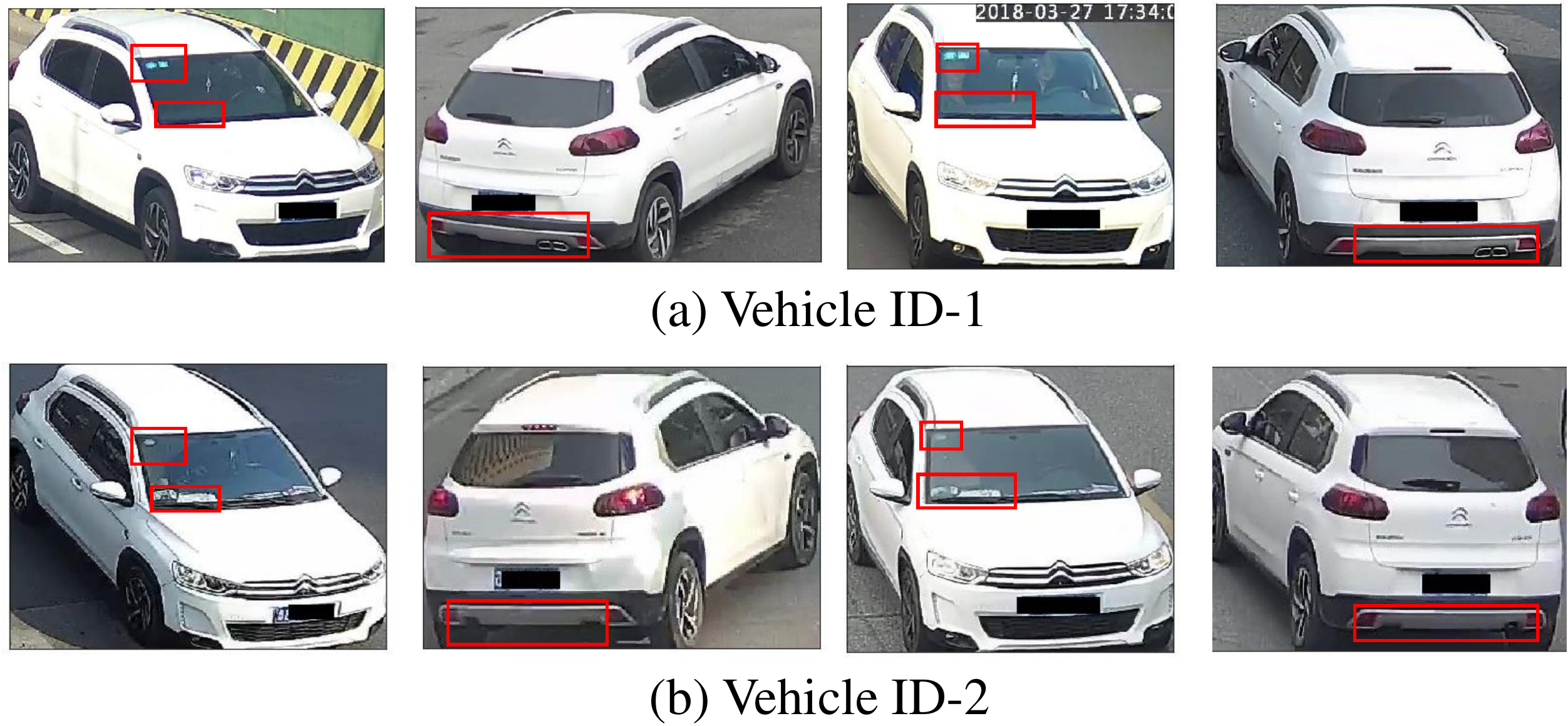}
    \caption{
    Toy examples from two different vehicles with the same type and color in VERI-Wild.
    Each row indicates different views of the same vehicle, which shows the challenge of large intra-instance difference.
    Each column denotes the same view from different vehicles, which shows the challenge of subtle inter-instance discrepancy.
    The red boxes represent the subtle discriminative differences of the two vehicles.
    }
    \label{fig:intro}
\end{figure}

    Vehicle Re-identification (ReID) has attracted more and more attention in recent years as it is important for building intelligent transportation and city surveillance systems~\cite{provid,vehicleid,veriwild,veri776, liu2019adaptive_xj, yang2019skeletonnet, coarse-to-fine}. 
This task aims to retrieve images of a query vehicle in a large gallery set, where the target vehicles are usually under various views and from widespread cameras.
It is particularly useful when the license plates of vehicles are occluded, blurred, and damaged.
As illustrated in Figure~\ref{fig:intro}, there exists two key challenges in this task,
1) the large intra-instance difference of the same vehicle under different views.
2) the subtle inter-instance discrepancy of different vehicles when they share the same type and color.

To address the above challenges, some works use the meta information (\eg vehicle attributes, spatial-temporal information) to improve the representation ability of the features.
Liu \etal~\cite{provid} proposed a course-to-fine search framework to model the attributes and spatial-temporal information into vehicle ReID.
Zheng \etal~\cite{zheng2019attributes} introduced a deep network to fuse the camera views, vehicle types and color into the features of vehicle.
These approaches focus on learning global representation for the vehicle.

However, the overall appearance changes dramatically under different view-points,
which results in the instability of global features and
also brings the first challenge.
In contrast, local features usually provide the stable discriminative cues.
Recently, researchers introduced local regions to learn the more discriminative features about the vehicle.
Wang \etal~\cite{sensetime17} generated orientation invariant features based on vehicle keypoints detection.
Liu \etal~\cite{ram} extracted local features based on three evenly separated regions of a vehicle to acquire distinctive visual cues.
He \etal~\cite{buaa19} detected window, lights, and brand for each vehicle through a YOLO detector to generate discriminative features.
The above methods focus on pre-defined regions to learn the subtle local cues.
However, as shown in Figure~\ref{fig:intro},
the distinctive cues (\eg exhaust, stickers and ornaments) may appear in any part of vehicle
and this leads to the second challenge.
\begin{figure*}[t]
    \centering
    \includegraphics[width=0.98\textwidth]{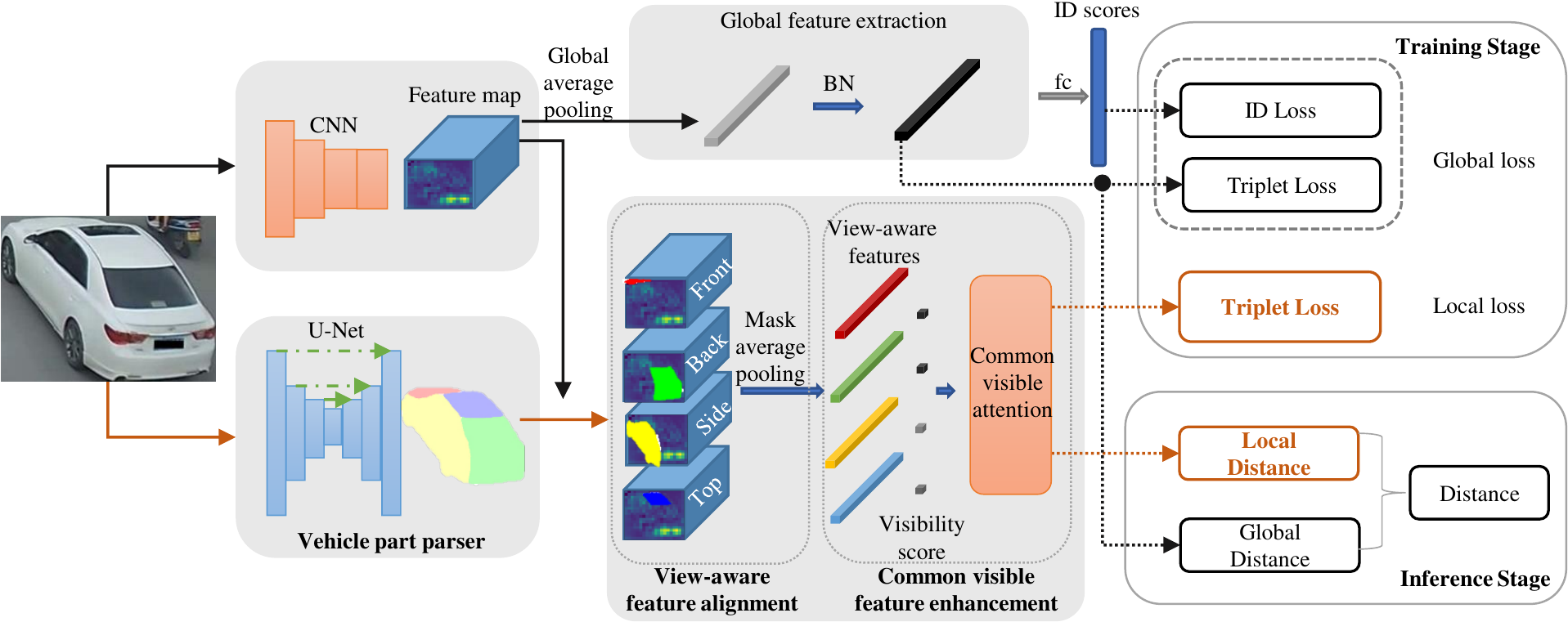}
    \caption{
    The network architecture of PVEN.
    First, the image is fed into feature extractor and vehicle part parser.
    The former outputs semantic feature maps while the latter generates the view mask of front, back, top and side.
    Then global feature of vehicle is extracted to construct the ID loss and triplet loss.
    View-aware features are extracted by mask average pooling for each mask.
    We aggregate the features by common-visible attention to formulate the triplet loss of local features.
    In inference stage, the distance of global feature and local features are added to get the final distance.
    }
    \label{fig:model}
\end{figure*}

Recently, data augmentation such as complementary views generation was applied to shrink the intra-instances discrepancy.
Zhou \etal~\cite{viewpoint_aware} tried to handle the multi-view problem based on generating the invisible views.
The generated views are derived from the visible view,
which are unable to reconstruct additional discriminative features.

In vehicle ReID, different views usually present the different characteristics of a vehicle.
We would acquire more discriminative description of a vehicle by leveraging these complementary characteristics.
However, since the same vehicle has large appearance discrepancy between different views,
how to effectively fuse such different characteristics remains a challenging problem.

To tackle the above challenge, this paper proposes a \textit{Parsing-based View-aware Embedding Network (PVEN)} to achieve the view-aware feature alignment and enhancement for vehicle ReID.
The PVEN consists of three modules: vehicle part parser, view-aware feature alignment, and common-visible feature enhancement.
First, we generate four view masks (front, back, top and side) by training a U-shape parsing network as shown in Figure~\ref{fig:segment_sample}.
Because the vehicle is a rigid body, the parsing network achieves an impressive accuracy as it need not handle the deforming problem.
Second, based on global feature maps, local view-aware features are aligned through mask average pooling.
Such alignment brings the vehicle the fine-grained representation with a complete spatial covering.
Third, we propose a common-visible attention to enhance the local features.
The mechanism tends to enlarge the effect of common visible views between two vehicles and suppress the non-salient views.
This helps to overcome the large intra-instance difference under different views and the subtle discrepancy of inter-instances under similar type and color.
Based on common-visible attention, we modified the typical triplet loss to avoid the mismatch of local features.
We optimize this local triplet loss and the global loss to learn the view-aware feature embedding.
As a result,
the global semantic and local subtle discriminative cues are jointly learned into the final embedding of the vehicle.


In summary, our main contributions are three folds.
\begin{itemize}
    \item To address the two key challenges in vehicles ReID, we propose a view-aware feature embedding method, where
    both feature alignment and enhancement of common visible views help to learn more robust and discriminative features.
    \item We introduce a common-visible attention to enhance features under different views.
    This not only shortens the distance among intra-instances, but also enlarges the discrepancy of inter-instances.
    \item Experiments on three vehicle ReID datasets verify the effectiveness of PVEN\footnote{https://github.com/silverbulletmdc/PVEN}.
    It achieves superior performance over SOTA methods with a large margin. 
\end{itemize}

    \section{Related Works}\label{sec:related-works}
Vehichle Re-identification has become a hot topic recently due to its wide using in intelligent transportation systems~\cite{provid,vehicleid,veriwild,veri776,coarse-to-fine,ealn,li2012learning}.
In previous works of vehicle ReID, these methods can be summarized into three groups:

(1)  \textit{Vehicle meta-information based feature fusion}.
The meta information, such as spatial-temporal information, vehicle attribute, are aggregated into global vehicle embeddings.
Liu \etal~\cite{provid} used a course-to-fine progressive search to leverage the vehicle attributes and spatial-temporal information.
Shen \etal~\cite{spatio_temporal_path} considered the constraint of spatial temporal information and used visual-spatial-temporal path to reduce searching space. 
Guided by camera views, vehicle types and color, Zheng \etal~\cite{zheng2019attributes} introduced a deep model to fuse the features for vehicle ReID.
These approaches learn global representation for vehicle, and they are sensitive to dramatic changes of view.
So they suffer from the challenge of large intra-instance difference of the same vehicle under different views.
(2) \textit{Local region based vehicle feature learning}.
Besides global features, recent works take advantage of local features to improve the representation ability.
For example, Wang \etal~\cite{sensetime17} generated orientation invariant feature based on pre-defined keypoints detection.
He \etal~\cite{buaa19} used the local region (e.g., window, brand and light bounding box) to learn more discriminative regions.
This type of methods usually depends on pre-defined distinctive region or key-points.
They ignore the fact that the discriminative cues may appear in any region of vehicle,
and suffer from the challenge of subtle inter-instance discrepancy of similar vehicles.
(3) \textit{Generative Adversarial Network based feature alignment}.
With GAN thriving, some works have started to introduce GAN into vehicle ReID.
For instance, Zhou \etal~\cite{viewpoint_aware} handled the viewpoint problem by generating the opposite side features using a GAN.
Lou \etal~\cite{veriwild} proposed to generate the hard samples by introducing a GAN.
Due to the limitation of generation ability of existing GAN
and the insufficient adversarial samples,
there exists large gap between the generated features and reality features.

Vehicle re-identification is also related with person ReID task,
which aims to find target persons from various views in a large set of persons.
Recently, CNN-based features achieved great progress on person ReID~\cite{pcb,zheng2019pyramidal,person_parsing,strongreid,Zhao2017Spindle,wei2018,zhang2017learning, liu2019adaptive}.
Sun \etal~\cite{pcb} split the image with a uniform partition strategy and extract CNN features for each parts.
Zhao \etal~\cite{Zhao2017Spindle} decompose the person by human body region to acquire human pose information.
Wei \etal~\cite{wei2018} proposed harmonious attention CNN to jointly learn attention selection and feature representation.
The explosion of person ReID methods lightens the vehicle ReID task.


    \section{Methodology}\label{sec:methodology}
To address the challenges of large intra-instance difference and subtle inter-instance discrepancy in vehicle ReID,
we propose a Parsing-based View-aware Embedding Network (PVEN).
It consists of three modules: vehicle part parser, view-aware feature alignment, and common-visible feature enhancement.
The PVEN focuses on the view-aware feature learning,
where the alignment and enhancement of common visible regions helps learn more robust and discriminative features.

\subsection{Vehicle Part Parser}\label{subsec:view-aware-vehicle-parsing}
As one key challenge of vehicle ReID, view transformation under multiple cameras is unavoidable.
Invariant feature learning under different views is an important insight to improve the performance of vehicle ReID.
We notice that most vehicles have the following two characteristics.
First, the vehicle can be regarded as a cube, which can be divided into different parts by view.
Second, the vehicle is the rigid body, so there are no physical deformations.
The characteristics imply that accurate vehicle parsing masks are capable to be extracted.
With these parsing masks, we can align corresponding parts for different vehicles.

\begin{figure}[t]
    \centering
    \includegraphics[width=\columnwidth]{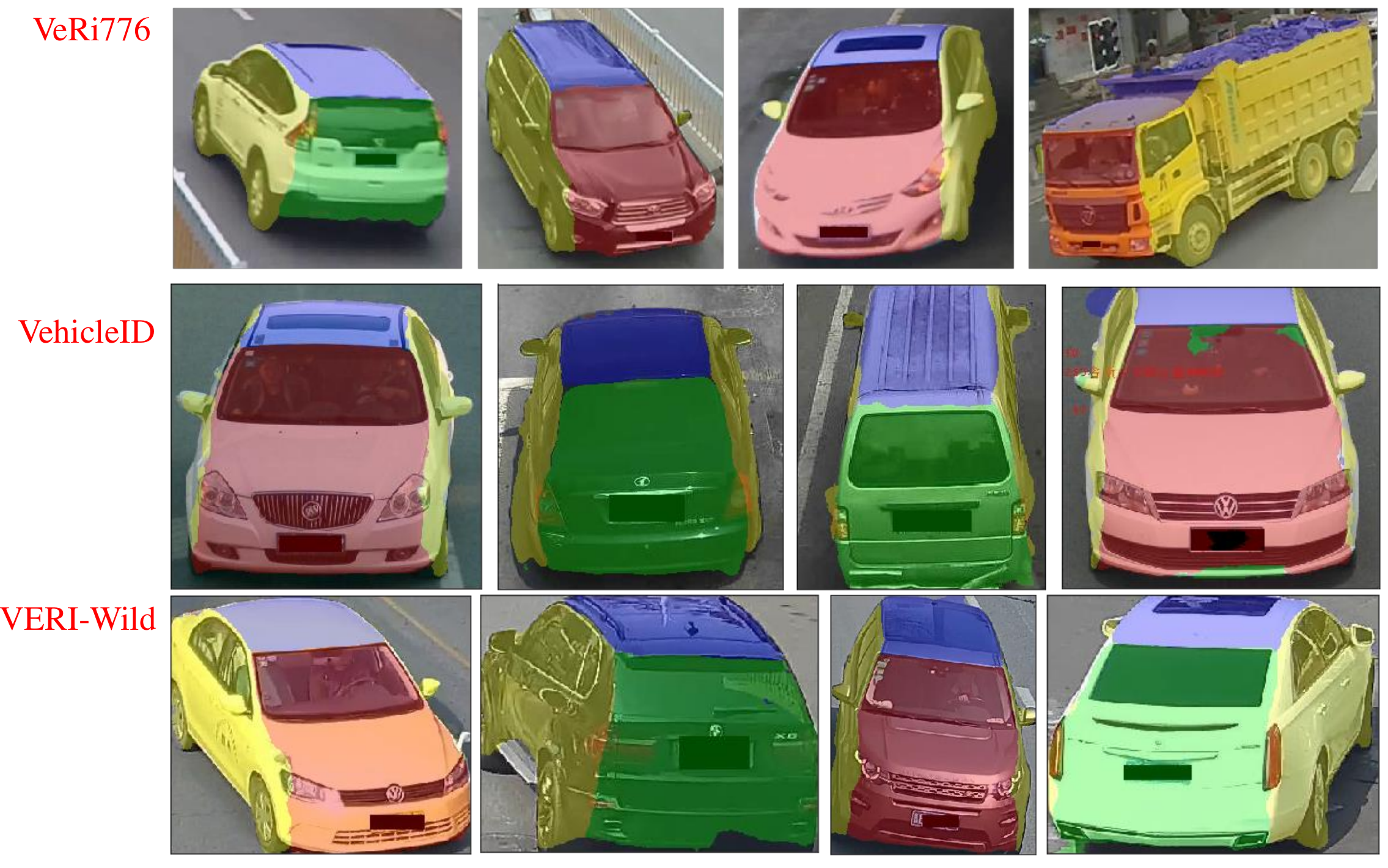}
    \caption{Examples of our parsing result on three main vehicle ReID datasets.
    The red, green, yellow and blue masks denote the front, back, side and top view of the vehicle respectively.
    }
    \label{fig:segment_sample}
\end{figure}

A vehicle can be roughly regarded as a cube with six surfaces.
The bottom of vehicle is usually invisible under the camera.
The left and right side of the vehicle usually can not appear at the same time under a certain view and are usually symmetry in visual.
Based on these observations, we parse a vehicle into four parts: front, back, side and top.
The side indicates the left or right side of a vehicle.
In this paper, the above parsing scheme is designed for vehicle view-aware representation.
As shown in Figure~\ref{fig:segment_sample}, there are two key advantages of this parsing scheme:
First, it covers the whole vehicle under the certain view,
so that every subtle differences between two vehicles can be captured.
Second, under most view-points, 3 parts of a vehicle are visible in an image,
which means that there are at least 2 same parts appearing in both the query and gallery image.



\textit{Parsing Annotation for VeRi776 Dataset.}
We annotate a subset of VeRi776~\cite{provid} dataset for training vehicle part parsing network.
To improve the adaptive capacity of the parsing model for various views,
we collect as many views of a vehicle as possible.
In detail, according the definition of viewpoint in ~\cite{sensetime17},
we sample images for seven different viewpoints of a vehicle.
If the number of viewpoints is less than four, we evenly sample four images of this vehicle.
Totally, we annotated 3165 images.
We select 2665 images of the annotated dataset randomly as training set and 500 images as validation set.

\textit{Vehicle Parsing Network.}
To get an accurate parsing result, we train a segmentation model~\cite{unet} using the above annotated dataset.
The parsing model takes SeResNeXt50~\cite{seresnext} as backbone and is trained with balanced cross entropy loss.
Our model achieves 81.2\% IoU score in the validation set,
which is sufficient for solving the view transformation challenge.
Figure~\ref{fig:segment_sample} shows some of the parsing results in three vehicle ReID datasets.
It shows the impressive generalization performance of the parsing model as the parser need not handle the deforming problem.

\subsection{View-aware Feature Alignment}\label{subsec:view-aware-feature-alignment}

Most of vehicle ReID models use deep global features to represent a vehicle,
which focus on learning high semantic information.
In this paper, we introduce the view-aware local features to obtain the fine-grained representation with a complete spatial covering.
Further, view-aware feature alignment is implemented to
avoid the mismatch among different views.

Here, we use ResNet50\cite{resnet} pre-trained on ImageNet\cite{imagenet} dataset
as our feature extractor.
We reset the stride of last pooling layer from 2 to 1
and obtain a $ 16 \times 16 \times 2048 $ feature map $\mathbf{F}$.
As shown in Figure~\ref{fig:model}, the feature extractor network has two output branches.
The first branch is the global branch, where
we apply the global average pooling to the feature map to get global feature $f_g$.
The other branch is the local branch for view-aware feature learning.
First, we pool the above view masks to $16 \times 16$
by max pooling, which is defined as
$\{ M_i|i\in \{1,2,3,4 \} \}$.
Second,
we apply the \textit{mask average pooling}
(MAP) to the feature map $F$ to compute four local
view-aware features $\{f_l^i|i\in\{0,1,2,3\}\}$.
They represent the front, back, side and top view
of a vehicle respectively.
The $f_l^i$ is calculated by
\begin{equation}
    f_{l}^i = \frac{\sum_{j,k=1}^{16} M_i(j, k) \times \mathbf{F}(j, k)}{\sum_{j,k=1}^{16} M_i(j,k)}
\end{equation}

The global feature blend features of different views into one feature.
It leads to a mismatch of views when comparing two vehicles.
Differently, the local view-aware features are aligned upon the above four views.
It decouples the information of different views into corresponding local features,
and provides view-aware embeddings for a vehicle.


\subsection{Common-visible Feature Enhancement}\label{subsec:view-aware-feature-aggregation}

\begin{figure}[t]
    \centering
    \includegraphics[width=\columnwidth]{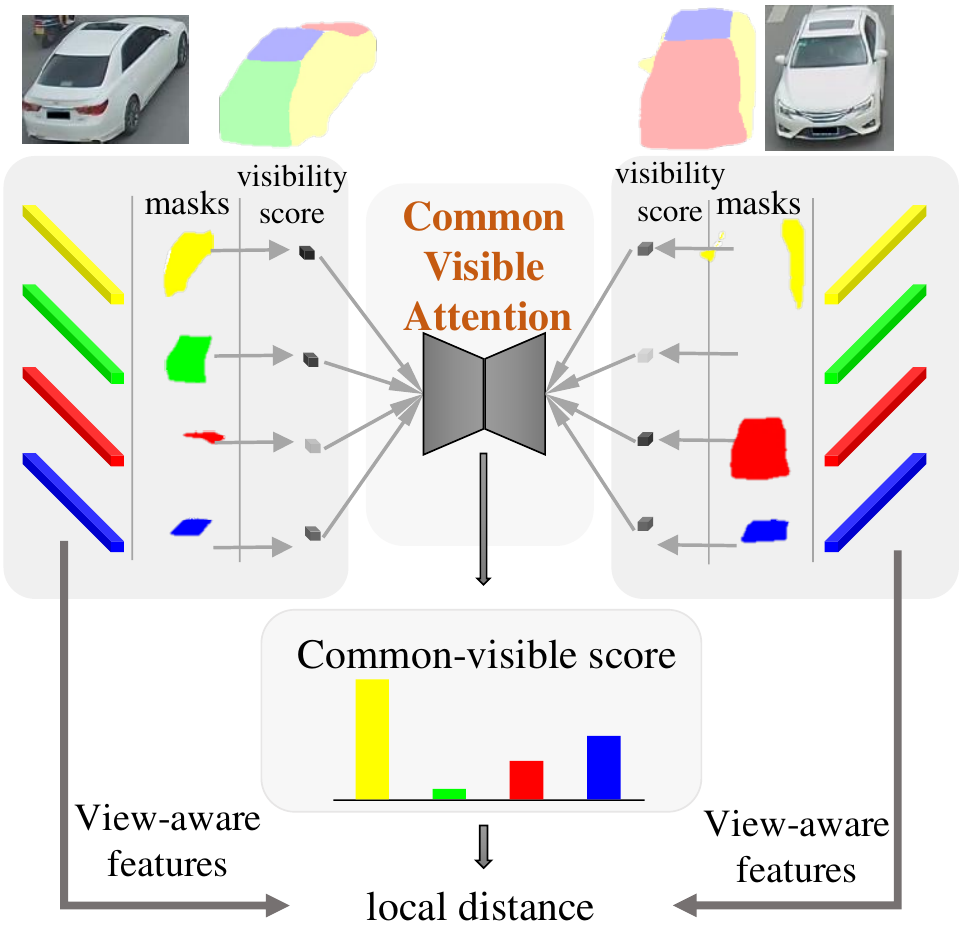}
    \caption{Illustration of common-visible attention.
    First, the visibility scores of different parts are computed based on the vehicle masks.
    Then, the common-visible scores of all parts are obtained by the common-visible attention.
    Finally, we calculate the local distance between two vehicles with their view-aware features and the corresponding common-visible scores.}
    \label{fig:loss}
\end{figure}

After the above stage, we obtain the four view-aware local features $f_l^i$ of the vehicle.
In this section, we introduce a common-visible attention to enhance the features of different views.
This helps capture the stable discriminative information of the same vehicle under different views.

Figure~\ref{fig:loss} shows the procedure of common-visible attention.
Given two image $p,q$, and their masks $M_i^p$ and $M_i^q$,
we compute the visibility score $v_i^p$ and $v_i^q$,
which indicated the size of corresponding area of each view.
The visibility score $v_i$ is defined as
\begin{equation}
    v_i=\sum_{j,k=1}^{16}M_i(j,k)
\end{equation}

We introduce the common-visible attention to compute the common-visible score ${a}^{p,q}_{i}$ as following,

\begin{equation}
{a}
    ^{p,q}_{i} = \frac{v_i^p v_i^q}{\sum_{i=1}^{N} v_i^p v_i^q}
\end{equation}
where ${a}^{p,q}_{i}$ measures the consistency of common visible regions.
Then, the distance of local features $\hat{D}$
between two vehicles is computed as,
\begin{equation}
    \label{eq:distance}
    \hat{D}^{p,q} = \sum_{i=1}^{N} a_i^{pq} D(f_i^p, f_i^q)
\end{equation}
where $D$ denotes the Euclidean distance.

If the vehicle lacks some views, the corresponding common-visible score would be relatively small.
So, only views with high score would contribute to the final distance.

In this paper, we optimize the network by constructing the loss of ID and triplet loss for global features, and triplet loss for local features.
The triplet loss of local features is calculated based on the above distance of local features as,
\begin{equation}
    L_{\text{triplet}}^l = \max(\hat{D}^{ap} - \hat{D}^{an} + \gamma, 0)
\end{equation}
where the local distance based on view-aware feature alignment and common-visible feature enhancement aims to reduce the intra-instance distance from different views and enlarge the inter-instance distance from similar vehicles.

Finally, the total objective of PVEN is to minimize the following loss,
\begin{equation}
    \label{eq:totalloss}
    L = L^g_{\text{id}} + L^g_{\text{triplet}} + L^l_{\text{triplet}}
\end{equation}

    \section{Experiments}\label{sec:experiments}
    \subsection{Datasets}\label{subsec:datasets}
We evaluate our model on three popular vehicle ReID datasets, including VehicleID~\cite{vehicleid}, VeRi776~\cite{provid} and VERI-Wild~\cite{veriwild}.

{\textbf{VehicleID}}~\cite{vehicleid} is a large scale vehicle ReID dataset.
It contains 221,763 images about 26,267 vehicles.
Images of this datasets are captured under front or back viewpoint.
Three test sets (i.e. small, medium and large) are extracted according to their size.
During inference stage, for each vehicle, one image is randomly selected as the gallery set while other images are regarded as query images.

{\textbf{VeRi776}}~\cite{provid} is also a classical vehicle ReID benchmark.
It consists of about 50,000 images of 776 vehicles,
which are collected by 20 cameras across block region under different viewpoints.
The training set covers 576 vehicles and the test set contains the other 200 vehicles.

{\textbf{VERI-Wild}}~\cite{veriwild} is another large scale dataset for vehicle ReID. It contains 416,314 images of 40,671 vehicles, which are collected by 174 cameras in a month.

%
%

\subsection{Experiments Setup}\label{subsec:training_and_inference}
\subsubsection{Training}
We train the parsing model for 40 epochs on our annotated Parsing VeRi dataset.
The batch size is 8 and the learning rate is 1e-4.
We use Adam as the optimizer.
Finally, The parser achieves 81.2\% IoU score in the validation set.

We train models for 120 epochs with warm-up strategy.
Initial learning rate is 3.5e-5, which increases to 3.5e-4 after the 10th epoch, and drops to 3.5e-5, 3.5e-6 in the 40th, 70th epoch for faster convergence.
We first pad 10 pixels on image border, and then randomly crop it to $256\times 256$.
We  also augment the data with random erasing.
Adam is used to optimize the model.
Further, we add a Batch Normalization layer after global feature.
A fully connected layer is added to map the global feature to ID classification score.


\subsubsection{Inference}
To evaluate our method, we first calculate the Euclidean distance $D_{global}$ among global features.
Then we calculate the distance $\hat{D}_{local}$ as defined in Eq.~\eqref{eq:distance} among local view-aware features.
Final distance between query set and gallery set is computed as
$\lambda_{1}D_{global}+\lambda_2 \hat{D}_{local}$.
Here, we set $\lambda_1 = 1$ and $\lambda_2 = 0.5$.

\subsubsection{Compared Methods}
We compare our method with some state-of-the-art methods, \eg
(1). \textit{Handcraft feature based methods.}
BOW-CN~\cite{zheng2015scalable} first adopts BOW model based on the Color Name (CN).
Local Maximal Occurrence Representation (LOMO)~\cite{liao2015person} is robust to the varied lightning conditions.
Fusion of Attributes and Color feaTures (FACT)~\cite{liu2016large} combines the low-level color feature and high-level semantic features.
(2). \textit{Deep learning based methods.}
GoogLeNet~\cite{yang2015large} is a GoogleNet\cite{googlenet} model fine-tuned on the CompCars~\cite{compcars} dataset.
Plate-SNN\cite{provid}, which use the number plate features to enhance the retrieval vehicles.
Siamese+Path~\cite{spatio_temporal_path} proposed the visual-spatial-temporal-path to exploit the temporal restrict.
GSTE~\cite{gste} proposed group-sensitive-triplet embedding to model the intraclass variance elegantly.
VAMI~\cite{viewpoint_aware} generated features of different views by GAN while
Feature Distance Adversarial Network~\cite{veriwild} (FDA-Net) generated the hard negative samples in feature space.
EALN~\cite{ealn} proposed an adversarial network that is capable of generating samples localized in the embedding space.
(3). \textit{Discriminitive region mining based methods.}
OIFE~\cite{sensetime17} used the 20 pre-defined keypoints to roughly align the vehicle features.
RAM~\cite{ram} split the image horizontally into 3 parts.
PRN~\cite{buaa19} detected the window, light and brand to capture the difference between vehicle instances.
AAVER~\cite{aaver} proposed an attention mechanism based on vehicle keypoints and orientation.

\subsection{Experiments on VehicleID dataset}\label{subsec:results-on-vehicleid}
\begin{table}[t]
    \centering
    \caption{The CMC@1 and CMC@5 on VehicleID.}
    \resizebox{\columnwidth}{!}{
    \begin{tabular}{c||c|c||c|c||c|c}
        \hline
        \multirow{2}{*}{Method} & \multicolumn{2}{c||}{small} & \multicolumn{2}{c||}{medium} & \multicolumn{2}{c}{large} \\ \cline{2-7}
        & @1 & @5 & @1 & @5 & @1 & @5         \\ \hline
        MD+CCL\cite{vehicleid} & 0.490 & 0.735 & 0.428 & 0.668 & 0.382 & 0.616       \\
        OIFE\cite{sensetime17} & - & - & - & - & 0.670 & 0.829       \\
        VAMI\cite{viewpoint_aware} & 0.631 & 0.833 & 0.529 & 0.751 & 0.473 & 0.703       \\
        RAM\cite{ram} & 0.752 & 0.915 & 0.723 & 0.870 & 0.677 & 0.845 \\
        EALN\cite{ealn} & 0.751 & 0.881 & 0.718 & 0.839 & 0.693 & 0.814     \\
        AAVER\cite{aaver} & 0.747 & 0.938 & 0.686 & 0.900 & 0.635 & 0.856     \\
        PRN\cite{buaa19} & 0.784 & 0.923 & 0.750 & 0.883 & 0.742 & 0.864       \\
        \hline
        PVEN & \textbf{0.847} & \textbf{0.970} & \textbf{0.806} & \textbf{0.945} & \textbf{0.778} & \textbf{0.920} \\ \hline
    \end{tabular}
    }
    \label{table:vehicleid_sota}
\end{table}

We compare the CMC@1 and CMC@5 scores on this dataset,
as there is only one ground-truth for each query vehicle.
Table~\ref{table:vehicleid_sota} shows the comparison results on three test datasets with different sizes.
We observe that, first, compared with other methods, the PRN and our PVEN obtained the performance improvement with a large margin.
This is because these two methods introduced the further learning to some key regions. This plays important role in vehicle ReID task.
Second, our PVEN achieve the improvement at the CMC@1 by $3.6\%+$ and CMC@5 by $4.5\%+$ over the SOTA PRN~\cite{buaa19} on different test data.
Although the PRN~\cite{buaa19}  introduced the detection about window, light, and brand for each vehicle, they ignored the fact that the distinctive cues will appear in any part of vehicle.
On the contrast, our method certifies the complete information mining of vehicle through the local view-aware feature embedding.
The above comparison results prove the effectiveness of the PEVN.

It is worth noting that the vehicle only contains two view-points in this dataset, namely, the front side and back side.
The extracted features from different views are completely different, even they are from the same vehicle.
Benefiting from view-aware feature enhancement, the PVEN can avoid the mismatch of local features under the different views.

\subsection{Experiments on VeRi776 dataset}\label{subsec:results-on-veri776}

We also evaluate the vehicle ReID methods on VeRi776 dataset, where three measurement metrics, including mAP, CMC@1 and CMC@5, are adopted.

Table~\ref{table:veri776_sota} shows the performance comparison among PVEN and other methods.
We find that, benefitting from learning of extra key regions,
both the PRN and our PVEN achieve a large promotion with a 16.0\% and 21.2\% on the mAP respectively.
Besides, different from the pre-defined regions of the PRN, the PVEN extracts the local information from four views,
which completely cover the whole vehicle. Thus, PVEN can learn the key distinctive and local cues to determine the target vehicle.
In detail, the PEVN obtains the improvement of 5.2\% on mAP, and 1.3\% CMC@1 over the PRN.
Moreover the CMC@5 of both methods have exceeded the 98.4\%, which is a promising performance for real vehicle ReID scenario.

\subsection{Experiments on VERI-Wild dataset}\label{subsec:experiments-on-veri-wild-dataset}

VERI-Wild dataset\cite{veriwild} is the current largest vehicle ReID dataset.
Here we compare our PEVN with other methods at three metrics, namely, CMC@1, CMC@5 and mAP.

Table~\ref{table:veriwild_map_sota} shows the performance of mAP on the three different size of test dataset.
We can find that our PVEN has a large promotion over the previous works of vehicle ReID.
In detail, the improvement of mAP is 47.4\%, 47.2\%, and 46.9\% on the small, medium and large dataset respectively.
This impressive boost of mAP benefits from the view-aware feature alignment and enhancement,
which help to learn more robust and discriminative features of vehicles.

\begin{table}
    \centering
    \caption{The mAP, CMC@1 and CMC@5 on VeRi776.}
    \resizebox{0.9\columnwidth}{!}{
    \begin{tabular}{@{}lccc@{}}
        \toprule
        Method & \multicolumn{1}{l}{mAP} & \multicolumn{1}{l}{CMC@1} & \multicolumn{1}{l}{CMC@5} \\ \midrule
        BOW-CN\cite{zheng2015scalable} & 0.122 & 0.339 & 0.537                     \\
        LOMO\cite{liao2015person} & 0.096 & 0.253 & 0.465                     \\
        GoogLeNet\cite{yang2015large} & 0.170 & 0.498 & 0.712                     \\
        FACT\cite{liu2016large} & 0.185 & 0.510 & 0.735                     \\
        FACT+Plate+STR\cite{provid} & 0.278 & 0.614 & 0.788                     \\
        Siamese+Path\cite{spatio_temporal_path} & 0.583 & 0.835 & 0.900                     \\
        OIFE\cite{sensetime17} & 0.480 & 0.894 & -                         \\
        VAMI\cite{viewpoint_aware} & 0.501 & - & -                         \\
        RAM\cite{ram} & 0.615 & 0.886 & 0.940 \\
        EALN\cite{ealn} & 0.574 & 0.844 & 0.941 \\
        AAVER\cite{aaver} & 0.612 & 0.890 & 0.947 \\
        PRN\cite{buaa19} & 0.743 & 0.943 & \textbf{0.989}                     \\
        \midrule

        PVEN & \textbf{0.795} & \textbf{0.956} & 0.984            \\ \bottomrule
    \end{tabular}
    \label{table:veri776_sota}
    }
\end{table}

\begin{table}[t]
    \centering
    \caption{The mAP on VERI-Wild.}
    \resizebox{0.9\columnwidth}{!}{
    \begin{tabular}{cccc}
        \toprule
        Method & small & medium & large \\ \midrule
        GoogLeNet\cite{yang2015large} & 0.243 & 0.242 & 0.215 \\
        Triplet\cite{schroff2015facenet} & 0.157 & 0.133 & 0.099 \\
        Softmax\cite{provid} & 0.264 & 0.227 & 0.176 \\
        CCL\cite{vehicleid} & 0.225 & 0.193 & 0.148 \\
        HDC\cite{hdc} & 0.291 & 0.248 & 0.183 \\
        GSTE\cite{gste} & 0.314 & 0.262 & 0.195 \\
        Unlable-GAN\cite{zhu2017unpaired} & 0.299 & 0.247 & 0.182 \\
        FDA-Net\cite{veriwild} & 0.351 & 0.298 & 0.228 \\ \midrule
        PVEN & \textbf{0.825}    & \textbf{0.770} & \textbf{0.697} \\ \bottomrule
    \end{tabular}
    }
    \vspace{-0.3cm}
    \label{table:veriwild_map_sota}
\end{table}

Table~\ref{table:veriwild_sota} shows the performance of CMC@1 and CMC@5 from different methods on three test datasets.
We can observe that, first, our PVEN exceeds all the other models under both metrics on different test datas.
The CMC@1 of the PVEN has the improvement of 32.7\%+ than the FDA-Net~\cite{veriwild}, and the CMC@5 of PVEN has the improvement of 16.4\%+ than FDA-Net.
The consistency of CMC promotion proves the effectiveness of our model.
Second, as the size of test datas increases, the performance of the traditional methods decreases with a large margin.
For example, for CMC@5, the state-of-the-art method FDA-Net declines 4.5\% between small and medium test data and 7.8\% between medium and large test data; the performance of PVEN degrades 0.4\% between small and medium test data and 1.0\% between medium and large test data;
This indicates that our approach has the better generalization ability under large datas.
This results from the view-aware features enhancement under different views in PVEN,
not only shortens the distance among intra-instances,
but also enlarges the discrepancy of inter-instances.

\begin{table}[tb]
    \centering
    \caption{The CMC@1 and CMC@5 on VERI-Wild.}
    \resizebox{\columnwidth}{!}{
    \begin{tabular}{c||c|c||c|c||c|c}
        \hline
        \multirow{2}{*}{Method} & \multicolumn{2}{c||}{small} & \multicolumn{2}{c||}{medium} & \multicolumn{2}{c}{large} \\ \cline{2-7}
        & @1 & @5 & @1 & @5 & @1 & @5         \\ \hline
        GoogLeNet\cite{yang2015large} & 0.572 & 0.751 & 0.532 & 0.711 & 0.446 & 0.636 \\
        Triplet\cite{schroff2015facenet} & 0.447 & 0.633 & 0.403 & 0.590 & 0.335 & 0.514 \\
        Softmax\cite{provid} & 0.534 & 0.750 & 0.462 & 0.699 & 0.379 & 0.599 \\
        CCL\cite{vehicleid} & 0.570 & 0.750 & 0.519 & 0.710 & 0.446 & 0.610 \\
        HDC\cite{hdc} & 0.571 & 0.789 & 0.496 & 0.723 & 0.440 & 0.649 \\
        GSTE\cite{gste} & 0.605 & 0.801 & 0.521 & 0.749 & 0.454 & 0.665 \\
        Unlabled Gan\cite{zhu2017unpaired} & 0.581 & 0.796 & 0.516 & 0.744 & 0.436 & 0.655 \\
        FDA-Net\cite{veriwild} & 0.640 & 0.828 & 0.578 & 0.783 & 0.494 & 0.705 \\ \hline
        PVEN & \textbf{0.967} & \textbf{0.992} & \textbf{0.954} & \textbf{0.988} & \textbf{0.934} & \textbf{0.978} \\ \hline
    \end{tabular}
    }
    \label{table:veriwild_sota}
\end{table}

\begin{table}[t]
    \centering
    \caption{Ablation study about each part of PVEN on VehicleID.}\smallskip
    \smallskip
    \begin{tabular}{@{}cccc@{}}
        \toprule
        settings & CMC@1 & CMC@5 & CMC@10          \\ \midrule
        PVEN w/o local & 0.796 & 0.937 & 0.969           \\
        PVEN w/o CV-ATT & 0.766 & 0.900 & 0.953          \\
        PVEN & \textbf{0.847} & \textbf{0.970} & \textbf{0.987}  \\ \bottomrule
    \end{tabular}
    \vspace{-0.3cm}
    \label{table:ablation_study}
\end{table}

\subsection{Ablation Study}



\subsubsection{The effectiveness of the parsing module}
To validate the effectiveness of the parsing model for vehicle ReID, we conduct an experiment that just evenly split the images vertically into four parts, and remain other settings the same with PVEN.
The results in Table~\ref{table:parsing} shows that parsing performs better than both the baseline and vertical split settings in mAP and CMC@5.

\subsubsection{The validation of view-aware feature learning}

We conduct ablation study about the proposed view-aware feature learning on VehicleID dataset.
PVEN w/o local indicates the PVEN model without the local branch of view-aware feature learning.
PVEN w/o CV-ATT adds the local branch, but do not use the common-visible attention.
It calculates the Euclidean distance of each local features.
Typical triplet loss are applied to the distance.
PVEN uses the full architecture as described in Section~\ref{sec:methodology}.
As in Table~\ref{table:ablation_study},
first, we observe that our PVEN achieves better accuracy than others by a large margin.
This is because view-aware feature alignment and common-visible attention drives the network
attending to the common visible parts between two compared vehicles.
Second, directly applying triplet loss to view-aware features without common-visible attention is harmful to the performance. It treats features of each view equally and ignore that features are non-salient under certain views,
so this introduces noise to the network.

%


\begin{table}[t]
    \centering
    \caption{The validation of parsing module on VeRi776.}\smallskip
    \smallskip
    \begin{tabular}{@{}cccc@{}}
        \toprule
        settings  & mAP  & CMC@1 & CMC@5  \\ \midrule
        baseline & 0.772 & 0.957 & 0.980 \\
        vertical split & 0.775 & 0.948 & 0.974 \\
        parsing & \textbf{0.795} & 0.956 & \textbf{0.984} \\ \bottomrule
    \end{tabular}
    \label{table:parsing}
\end{table}

\begin{table}[t]
    \centering
    \caption{Weight selection of global and local distance on VeRi776.}\smallskip
    \smallskip
    \begin{tabular}{@{}cccccc@{}}
        \toprule
        $\lambda_1$ & $\lambda_2$ & mAP & CMC@1 & CMC@5 & CMC@10          \\ \midrule
        1 & 0 & 0.787 & 0.955 & 0.982 & 0.990          \\
        1 & 0.3 & 0.794 & 0.956 & 0.984 & 0.992          \\
        1 & 0.5 & \textbf{0.795} & 0.956 & \textbf{0.984} & \textbf{0.992} \\
        1 & 0.7 & 0.794 & 0.959 & 0.984 & 0.992          \\
        1 & 1 & 0.793 & \textbf{0.961} & 0.984 & 0.991          \\
        0 & 1 & 0.713 & 0.922 & 0.967 & 0.982          \\ \bottomrule
    \end{tabular}
    \vspace{-0.3cm}
    \label{table:lambda}
\end{table}

\subsubsection{Weight selection of global and local distance}
\label{subsec:discussion-on-local-and-global-distance}
Here we conduct experiments to figure out how the view-aware feature affects the performance of vehicle ReID.
Table~\ref{table:lambda} shows the results of different weights between global and local distance.
We can find that the view-aware local feature brings the improvement of final results at all  metrics, namely mAP, CMC@1, CMC@5, CMC@10.
The local view-aware feature learning helps the global features learn better.



\subsubsection{Visualization of view-aware feature learning}
To better understand the influence of view-aware feature learning in PVEN,
we visualize the distance heatmap of vehicle images.
The pixels with high score in distance heatmap indicate that they play more important role in determine the similarity between query and gallery vehicle.
Specifically, the heatmap is the weighted sum of the last feature maps of the backbone.
The weights are computed from the element-wise Euclidean distance of two features.

\begin{figure}[t]
    \centering
    \includegraphics[width=\columnwidth]{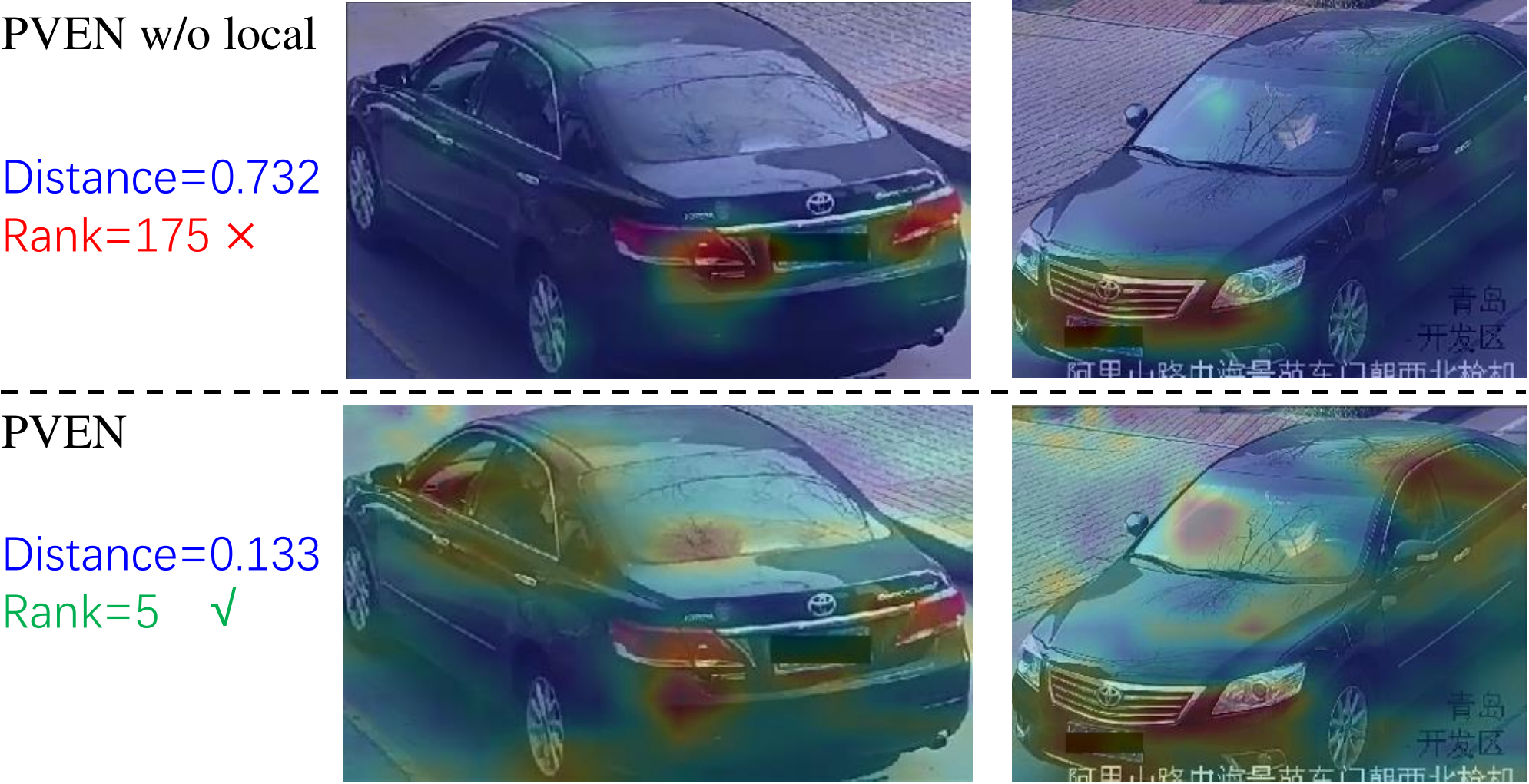}
    \caption{The distance heatmap of two images for the same vehicle.
    These two images are under different views (back and front).
    The PVEN without view-aware future learning (first row) mainly focuses on the back and front part of the vehicle,
    while PVEN (second row) pays more attention on the common visible parts.}
    \label{fig:heatmap}
    \vspace{-0.3cm}
\end{figure}

Figure~\ref{fig:heatmap} shows the distance heatmap of two images from our PVEN and PVEN without view-aware feature learning.
The two images are from back view and front view respectively, but they are of the same vehicle.
As shown the first row in Figure~\ref{fig:heatmap},
we observe that the PVEN without view-aware feature learning mainly focus on the mismatched back and front lights of the vehicle.
As a comparison in the second row,
the PVEN pays more attention to the co-occurrence regions, such as the side and top part of the vehicle,
which decreases the obstruction of different views and shortens the distance among intra-instances.


\subsection{Cross-dataset Transferring}

In vehicle Re-ID task, the main difference between different datasets is the distribution of views.
For example, most vehicles are under back and front views in VehicleID while more vehicles are under side views in VeRi776 and VERI-Wild.
We test the transferring performance of PVEN to evaluate the effectiveness of the view-aware feature embedding.
We train the PVEN in VERI-Wild and test it in the VehicleID.
The compared methods RAM~\cite{ram} and EALN~\cite{ealn} are trained and tested both on VehicleID.

Table~\ref{table:cross_domain} shows the performance.
We found the performance of our model outperforms RAM~\cite{ram} and EALN~\cite{ealn} on CMC@1 and CMC@5,
although PVEN is not trained on the test dataset.
The proposed view-aware feature learning alleviates the multi-view distribution difference.
This transfer ability of PVEN derives from two aspects.
First, the view-aware feature alignment brings the fine-grained representation for the vehicle.
Second, the common-visible feature enhancement under different views can
not only shorten the distance among intra-instances, but also enlarge the discrepancy of inter-instances.

\subsection{Qualitative Analysis}\label{subsec:qualitative-analysis}

\begin{table}[b]
    \centering
     \vspace{-0.1cm}
    \caption{The mAP, CMC@1 and CMC@5 on cross domain setting.}\smallskip
    \resizebox{.95\columnwidth}{!}{
    \smallskip
    \begin{tabular}{@{}ccccc@{}}
        \toprule
        method & train & test & CMC@1 & CMC@5          \\ \midrule
        RAM\cite{ram} & VehicleID & VehicleID & 0.752 & 0.915 \\
        EALN\cite{ealn} & VehicleID & VehicleID & 0.751 & 0.881 \\ \hline
        PVEN & VERI-Wild & VehicleID & 0.772 & 0.944          \\ \bottomrule
    \end{tabular}
    }
    \label{table:cross_domain}
\end{table}

\begin{figure}[t]
    \centering
    \includegraphics[width=\columnwidth]{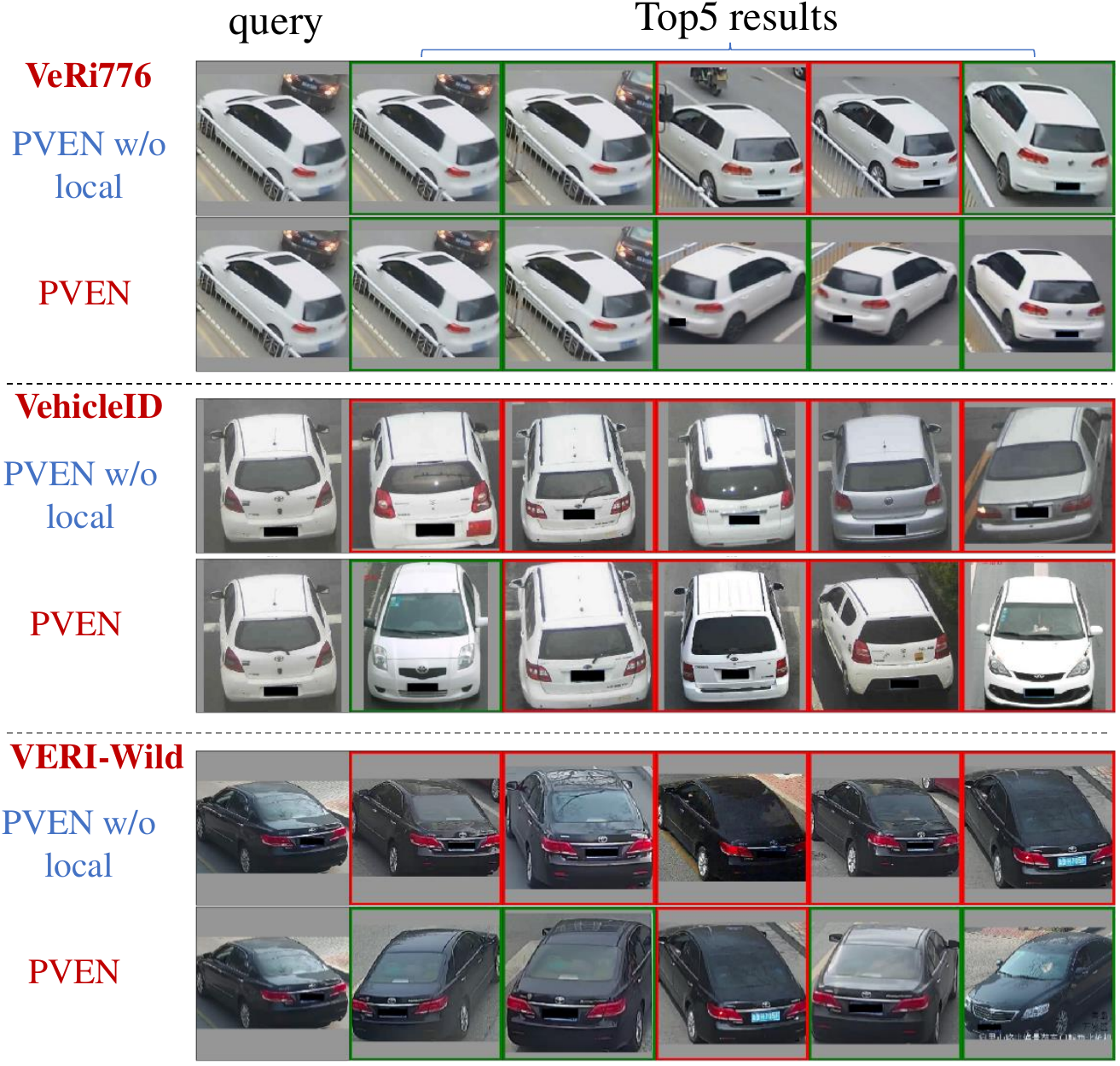}
    \caption{Visualization of ranking list on vehicle ReID task.
    The images in the first column are the query images.
    The rest images are retrieved top-5 ranking results.
    The correct retrieved images are in green border,
    while false instances are in red border.}
    \label{fig:rank_list}
    \vspace{-0.3cm}
\end{figure}

Figure~\ref{fig:rank_list} shows the qualitative results of our PVEN on the three vehicle ReID datasets.
We can observe that when the query and target images are under different views,
our PVEN can better recognize the same vehicle, which benefits from view-aware feature alignment and enhancement.

In Figure~\ref{fig:rank_list}, The top two rows show the results on VeRi776.
The medium two rows are the results on VehicleID, where for each test query, there is only one target image in the gallery set.
The bottom two rows show the results on VERI-Wild.
We can find that the top-k retrieved vehicles of the PVEN without view-aware feature learning are of the same view,
and also in similar color and vehicle type.
As comparison, PVEN retrieves the right images under different views.
This indicates that the view-aware feature learning can better align and enhance
the local features so as to ease the view transformation problem.

    \section{Conclusion}\label{sec:conclusion}
    In this paper, we propose a parsing-based view-aware embedding network.
With the help of vehicle part parser,  view-aware feature alignment provides a fine-grained representation of the vehicle.
We design a common-visible attention to focus on the common visible views, which enhance the vehicle embeddings under different views.
This not only shortens the distance among intra-instances, but also enlarges the discrepancy of inter-instances.
PVEN helps capture the stable and discriminative information of the same vehicle.
The experiments on three datasets show that our model outperforms SOTA methods by a large margin.

    \textbf{Acknowledgement.} This work was supported in part by National Key R\&D Program of China under Grand: 2018AAA0102003, in part by National Natural Science Foundation of China: 61771457, 61732007, 61772497, 61772494,  61931008, U1636214, 61622211, U19B2038, 61702491, in part by Key Research Program of Frontier Sciences, CAS: QYZDJ-SSW-SYS013.

    {\small
    \bibliographystyle{ieee_fullname}
    \bibliography{mybib}
    }
\end{document}